\documentclass[10pt,twocolumn,letterpaper]{article}

\usepackage{cvpr}              

\usepackage{graphicx}
\usepackage{amsmath}
\usepackage{amssymb}
\usepackage{booktabs}

\usepackage[pagebackref,breaklinks,colorlinks]{hyperref}

\usepackage[accsupp]{axessibility}  

\usepackage[capitalize]{cleveref}
\crefname{section}{Sec.}{Secs.}
\Crefname{section}{Section}{Sections}
\Crefname{table}{Table}{Tables}
\crefname{table}{Tab.}{Tabs.}

\def\cvprPaperID{18} 
\def\confName{CVPR}
\def\confYear{2022}

\begin{document}

\title{Coarse-to-Fine Cascaded Networks with
Smooth Predicting for \\ Video Facial Expression Recognition}

\author{Fanglei Xue$^{1,2}$\thanks{
This work was done when Fanglei Xue was an intern at Institute of Deep Learning, Baidu Research.}, Zichang Tan$^{3,4}$, Yu Zhu$^{3,4}$, Zhongsong Ma$^{1,2}$, Guodong Guo$^{3,4}$\thanks{Corresponding author}\\
$^1$University of Chinese Academy of Sciences, Beijing, China\\
$^2$Key Laboratory of Space Utilization, Technology and Engineering Center for Space Utilization, \\
Chinese Academy of Sciences, Beijing, China \\
$^3$Institute of Deep Learning, Baidu Research, Beijing, China\\
$^4$National Engineering Laboratory for Deep Learning Technology and Application, Beijing, China\\
{\tt\small xuefanglei19@mails.ucas.ac.cn mazhongsong@csu.ac.cn  \{tanzichang, zhuyu05, guoguodong01\}@baidu.com}
}

\maketitle

\begin{abstract}
Facial expression recognition plays an important role in human-computer interaction. In this paper, we propose the Coarse-to-Fine Cascaded network with Smooth Predicting (CFC-SP) to improve the performance of facial expression recognition. CFC-SP contains two core components, namely Coarse-to-Fine Cascaded networks (CFC) and Smooth Predicting (SP). For CFC, it first groups several similar emotions to form a rough category, and then employs a network to conduct a coarse but accurate classification. Later, an additional network for these grouped emotions is further used to obtain fine-grained predictions.  For SP, it improves the recognition capability of the model by capturing both universal and unique expression features. To be specific, the universal features denote the general characteristic of facial emotions within a period and the unique features denote the specific characteristic at this moment. Experiments on Aff-Wild2 show the effectiveness of the proposed CFSP. 
We achieved 3rd place in the Expression Classification Challenge of the 3rd Competition on Affective Behavior Analysis in-the-wild.
The code will be released at \url{https://github.com/BR-IDL/PaddleViT}.
\end{abstract}


\section{Introduction}
\label{sec:intro}
In the past years, facial expression recognition has attracted increasing attention in the field of computer vision and psychology science~\cite{xue2021TransFER,weng2021attentive,zhang2021prior,jin2021multi,li2017reliable,farzaneh2021facial,zeng2018facial,mollahosseini2017affectnet}. 
Facial expression can be regarded as the external signals to reflect people's psychological activities and intentions. Therefore, facial expression recognition is particularly important in human-computer interaction. Although this topic has been studied for many years, it is still difficult to achieve accurate emotional analysis nowadays.

In order to promote the development of facial expression recognition,
Kollias et al.~\cite{zafeiriou2017aff,kollias2020analysing,kollias2019expression,kollias2019deep,kollias2022abaw,kollias2021analysing,kollias2021distribution,kollias2021affect,kollias2019face} organize the competition on Affective Behavior Analysis in-the-wild (ABAW). The competition has been successfully held two times, in conjunction with IEEE FG 2020 and ICCV 2021. 
The third competition (ABAW3) will be held in conjunction with CVPR 2022.
Those competitions provide several large-scale in-the-wild benchmarks for affective behavior analysis,
e.g., Aff-Wild2~\cite{kollias2019expression}.
Aff-Wild2 contains 558 videos with around 2.8 million frames with rich annotations.
Besides the common annotations of facial expressions and Action Units (AUs),
it also provides the annotations of valence and arousal to
study continuous emotions.

\begin{figure*}[]
\begin{center}
\includegraphics[width=\linewidth]{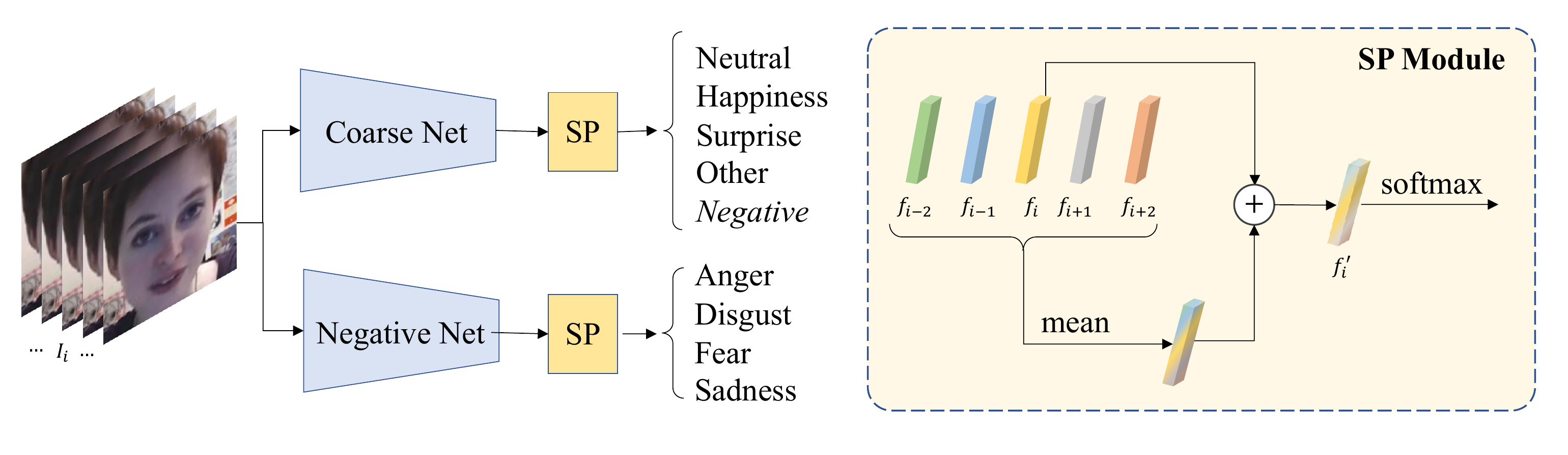}
\end{center}
\caption{The architecture of our model. Given the input frames $I_{i-2}$, ... $I_i$, ..., $I_{i+2}$, we first extract the features of these frames by the Coarse Net and get the coarse results. If the predicted coarse result is \textit{Negative}, the Negative Net is utilized to predict the negative classes. The smooth predicting (SP) module is designed to take multiple frame features (denote as $f_i$) as input. And every feature is calculated by adding themselves to the mean values of their adjacent features. After this, the predicted result will be more smooth and stable.}
\label{fig:model}
\end{figure*}

The goal of our work is to study facial expression recognition based on Aff-Wild2.
One main challenge in facial expression recognition is the problem of label ambiguity,
which may be caused by two aspects:
one is the ambiguity of the expression itself, where some expressions are similar and difficult to distinguish.
The other is the labeling ambiguity caused by different people having different understandings of different facial expressions, resulting in inconsistent labeling.
For example, 'Sadness' is very similar to 'Disgust',  and it is hard to distinguish them.
To address this problem, we propose the coarse-to-fine cascaded networks to obtain more reliable predictions
in a coarse-to-fine manner.
Specifically, we first group several similar emotions as a coarse category (including 'Anger', 
'Disgust', 'Fear' and 'Sadness'),
and this coarse category will be combined with other emotions to train a coarse model at the first stage.
In the second stage, an additional network is further employed to distinguish
those grouped emotions and then obtain fine predictions.
Moreover, inspired by the previous work~\cite{shi2021learning}, 
the feature embedding of facial expression can be divided into two parts, namely universal and unique parts.  The universal part describes the general characteristic of facial emotions, 
and the unique part describes the specific features of the expressive component.
The unique part can be denoted by the features of each specific image.
For the universal part, it is captured by using a proposed Smooth Predicting (SP),
which collects the features of nearby iterations and outputs the averaging features.
The final feature used for prediction is the sum of the universal and unique parts.
Furthermore, we train multiple models based on various architectures (e.g., ResNet~\cite{he2016deep}, Swin-Transformer~\cite{liu2021swin}) and improve the performance by using an ensemble of them.

The main contributions of our work can be summarized as follows:
\begin{itemize}
  \item We propose the Coarse-to-Fine Cascaded networks (CFC) to address the label ambiguity problem in facial expression recognition. 
  \item We propose a Smooth Predicting (SP) to capture both universal and unique features, which can effectively improve the performance of facial expression recognition.
\end{itemize}



\section{Related Works}
In this section, we briefly review the related works in
facial expression recognition and affective behavior analysis.

\textbf{Facial Expression Recognition.}
Facial expression recognition~\cite{xue2021TransFER,weng2021attentive,zhang2021prior,jin2021multi,li2017reliable,farzaneh2021facial,zeng2018facial,mollahosseini2017affectnet}, which is a classic research topic in computer vision and psychology science, has been studied for many years.
Most of the early methods were based on hand-crafted features~\cite{guo2005learning,zhao2007dynamic,zhao2011facial}.
Recently, deep learning has greatly advanced the development of facial expression recognition~\cite{xue2021TransFER,weng2021attentive,zhang2021prior,jin2021multi,farzaneh2021facial,zeng2018facial,She_2021_CVPR}.
For example, Weng et al.~\cite{weng2021attentive} propose a multi-branch network to capture both global and local features for facial expression recognition. Xue et al.~\cite{xue2021TransFER} propose a hybrid architecture with combined CNN and transformer,
which achieves state-of-the-art performance on multiple benchmarks.
Some researchers~\cite{wang2020region,li2018occlusion,li2020attention} propose to extract discriminative features 
by attention mechanism, which is proven to be robust for occlusions.

\textbf{Affective Behavior Analysis.}
This subsection mainly reviews the related works in previous competitions of ABAW.
Kuhnke et al.~\cite{kuhnke2020two} propose a two-stream network to utilize both vision and audio information 
for affective behavior analysis.
Considering the problem of incomplete labels, Deng et al.~\cite{deng2020multitask} propose to learn from unlabeled data by a teacher-student scheme. 
Moreover, Zhang et al.~\cite{zhang2021prior} a Prior Aided Streaming Network (PASN) for multi-task expression recognition, including Categorical Emotions (CE), Action Units (AU), and Valence Arousal (VA).
In PASN, three tasks are placed in a serial manner (AU$\rightarrow$CE$\rightarrow$VA),
where the basic up-streaming task would aid the recognition of down-streaming tasks.
Jin et al.~\cite{jin2021multi} propose a multi-modal multi-task method for facial action unit and expression Recognition.


\begin{figure*}[]
\begin{center}
\includegraphics[width=\linewidth]{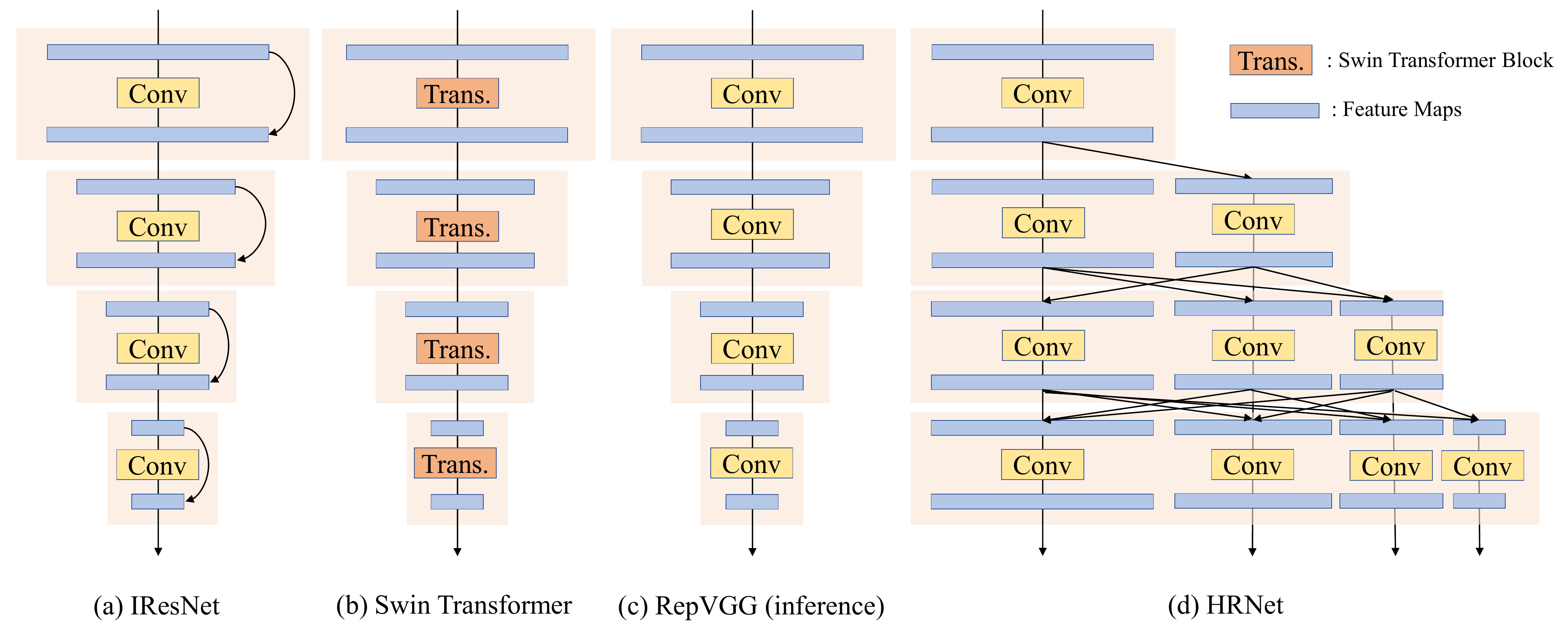}
\end{center}
\caption{Sketch of different backbone architectures. They have various feature extract methods (convolution or self-attention), various lateral connections (with or without skip connections), and various resolution streams (gradually decrease or keep the same). We believe different architecture could learn different features. Thus, we ensemble these four backbones in the coarse net to generate robust prediction results and prevent over-fitting.}
\label{fig:backbones}
\end{figure*}

\section{Method}

\subsection{Overview}
The overall architecture of our proposed method is illustrated in Fig.~\ref{fig:model}. It mainly consists of two parts: the Coarse-to-Fine Cascaded networks (CFC) on the left and the Smooth Predicting (SP) module on the right. In the following, we will introduce these two modules in detail.

\subsection{Coarse-to-Fine Cascaded Networks}
Facial expressions are generated by muscle movement in faces. Prior studies \cite{ekman1993facial} split expressions into seven basic categories to represent the common emotions shared by all human beings. The EXPR task in ABAW3 aims to categorize facial images into these seven expressions plus an "other" category. But these seven categories are not equally distributed, some expressions are very similar to others, especially negative expressions. Specifically, four negative expressions (anger, disgust, fear, and sadness) have very similar muscle movements in the same facial areas \cite{Ekman1978FacialAC}. Compared with other expressions, like happiness and surprise, these expressions are more difficult to predict correctly. On the other hand, most in-the-wild facial expression datasets are collected from the Internet, where most people share their happy times. Negative emotions are hard to obtain in real scenes \cite{li2021adaptively}, making existing FER datasets imbalance. To address these two issues, we adopted the coarse-fine strategy in \cite{li2021adaptively}.

To be specific, we utilize two stages to predict coarse and negative labels, separately. Following \cite{li2021adaptively}, we combine four negative expressions (anger, disgust, fear, and sadness) to one category (negative) in the first stage, resulting in five classes in the coarse stage. If the coarse net predicts the facial image into the negative class, the image will be fed into the negative net to predict the negative expression. Otherwise, the coarse label will be returned directly. In this manner, the low-frequency negative expression samples are combined in the negative, making the coarse dataset more balance. And the negative net only aims to distinguish negative expressions, which share a similar difficulty.

In order to further enhance the robustness of our framework, we adopt the model ensemble strategy to our coarse net. Specifically,  IResNet-152 \cite{deng2019arcface}, Swin-S \cite{liu2021swin}, RepVGG  \cite{dingRepVGGMakingVGGstyle2021} and HRNet \cite{wang2021DeepHighResolution} are adopted as backbone networks. As illustrated in Fig.~\ref{fig:backbones}, there are significant differences among these architectures.  IResNet (a), Swin Transformer (b), and RepVGG (c) utilize pooling or path merging layers to decrease the spatial resolution of feature maps to reduce the computation cost and enlarge the receptive field, while HRNet (d) remains the high-resolution representations and exchange features across resolutions to obtain rich semantic features. (b) utilizes the self-attention mechanism with shifted windows to explore relationships with different areas while the other three models adopt the traditional convolution method. (a) utilizes residual connection within stages while (b) and (c) do not (c) has skip-connection in the training stage. (d) adopted more connections among different resolutions to extract rich semantic features. These different architectural designs could help different models learn different features, and prevent the whole framework overfit to some noisy features.

\begin{figure}[]
\begin{center}
\includegraphics[width=\linewidth]{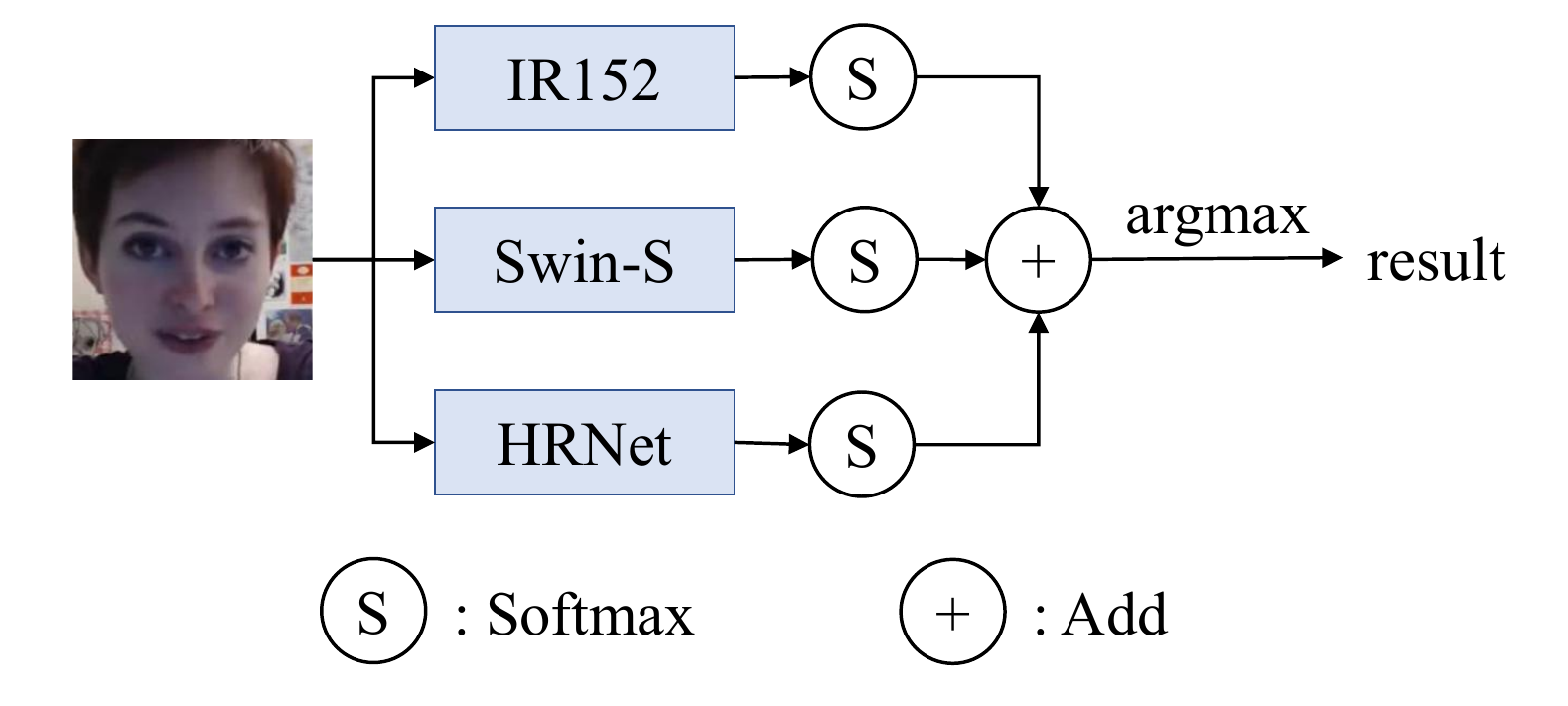}
\end{center}
\caption{Illustration of our ensemble process. Feature logits are extracted from each backbone network separately and produce the per-class confidential scores by the softmax function. The scores are then summed up and generate the final result with the argmax function.}
\label{fig:backbones}
\end{figure}

\subsection{Smooth Predicting}
Benefit from the timeline in videos, we can find the process of muscle movement, \eg the process from neural to happiness. So the expression change in videos should be very smooth. But current 2D FER methods only consider the spatial information in facial images and ignore the relative information among frames. They predict every frame independently, which may lead to different results in the same expression clip. Videos in the ABAW3 dataset have a high frame rate (30 fps), thus it does not make sense for expressions changing very frequently in adjacent frames. 

To address this issue, we propose a smooth predicting module to capture both universal and unique features of adjacent frames. As illustrated in the right part of Fig.~\ref{fig:model}, the SP module takes a series of feature logits as input and enhance every feature by adding the average information among a specific window (5 in Fig.~\ref{fig:model} as an example) to the original feature. 

Mathematically, given $n$ adjacent frame images, $I_0, I_1, ... I_{i-1}, I_i, I_{i+1},... I_{n-1}$, and the windows size ($w$), we first utilize the network to extract the features from every images independently. After this, $n$ image features will be obtained, named $f_0, ... f_i, ...f_{n-1}$. Then, our proposed SP module will update every frame feature by:

\begin{equation}
    f_{i}^{'} = f_i + \frac{\sum_{j=i-w/2}^{i+w/2}f_j}{ w}
\end{equation}

where the $f_{i}^{'}$ denotes the updated feature. The updated feature consists of two-part: the unique current frame feature and the universal feature among the given window size. We find that this will improve the smoothness of model outputs as well as the performance.

\begin{table}[]
\centering
\begin{tabular}{lcc}
\toprule
Methods             & F1 (\%)    \\ \midrule
Baseline            & 23 \\
Ours baseline       & 40  \\
Ours baseline + C-F & 42  \\
Ours baseline + SP  & 43  \\
Ours                & \textbf{46}  \\ \bottomrule
\end{tabular}
\caption{The ablation studies to our method with and without C-F strategy and SP module on the validation set of Aff-Wild2.}
\label{tab:ab}
\end{table}

\section{Experiments}
In this section, we will first introduce some training details and then give some experiment results based on the official validation set of the Aff-Wild2 \cite{kollias2019expression}. 

\subsection{Details}
We only used the Aff-Wild2 \cite{kollias2019deep} to train our models. All videos are processed into frame images by FFmpeg, and the MTCNN \cite{zhang2016joint} is utilized to detect and align facial images. All the facial images are cropped into 112$\times$112 except for Swin Transformer, which is 224 $\times$ 224. 
To alleviate the over-fitting problem, we utilized several data augmentation methods following \cite{xue2021TransFER}: random rotate and crop, random flip, colorjit, blur, and random erasing \cite{zhong2020random}. The over-sample method proposed in \cite{Gupta2019LVIS} is also adopted to reduce the influence of the class imbalance problem.

An ensemble of IR152 \cite{deng2019arcface}, Swin-S \cite{liu2021swin}, and HRNet \cite{wang2021DeepHighResolution} is adopted as our coarse net, and the RepVGG \cite{dingRepVGGMakingVGGstyle2021} is adopted as the negative nets. All these backbone nets are pre-trained on the Ms-Celeb-1M \cite{guo2016ms}.

The cross-entropy loss is applied for training. And the f1-score (\%) on the validation set is reported by default.
The detail of hyperparameters could be found on the corresponding config file in our released codes. All of the experiments are trained on one NVIDIA V100 graphic card.

\begin{table*}[]
\centering
\begin{tabular}{cccc|cc|c}
\toprule
\multicolumn{4}{c|}{Coarse}      & \multicolumn{2}{c}{Negative} &  \\  \midrule
Swin-S & IR152 & HRNet & RepVGG & Swin-S & RepVGG & F1 (\%)    \\ \midrule
$\checkmark$&       &       &        &        & $\checkmark$& 43.78 \\
       & $\checkmark$&       &        &        & $\checkmark$& 42.74 \\
       &       & $\checkmark$&        &        & $\checkmark$& 43.78 \\
    &       & &  $\checkmark$ &        & $\checkmark$& 40.38 \\
$\checkmark$& $\checkmark$&       &        &        & $\checkmark$& 44.25 \\
$\checkmark$& $\checkmark$&       &        & $\checkmark$&        & 43.28 \\
$\checkmark$& $\checkmark$&       &        & $\checkmark$& $\checkmark$& 43.17 \\
$\checkmark$&       & $\checkmark$&        &        & $\checkmark$& 43.28 \\
$\checkmark$& $\checkmark$& $\checkmark$&        &        & $\checkmark$& \textbf{46.15} \\
$\checkmark$& $\checkmark$& $\checkmark$&        & $\checkmark$& $\checkmark$& 44.62 \\ \bottomrule
\end{tabular}
\caption{Results of different ensemble strategies on the validation set of the Aff-Wild2. The Swin-S, IR152, and HRNet ensembled as the coarse net, and the RepVGG as the negative net performs the best.}
\label{tab:ensemble}
\end{table*}

\subsection{Results}

\begin{table}[]
\centering
\begin{tabular}{ccc}
\toprule
Backbone & Coarse & Negative \\ \midrule
IR152  \cite{deng2019arcface}   & 50.82  & 44.99    \\
Swin-S \cite{liu2021swin}   & \textbf{51.52}  & 46.70    \\
HRNet \cite{wang2021DeepHighResolution}    & 50.67  & 45.60    \\
RepVGG \cite{dingRepVGGMakingVGGstyle2021}   & 48.40  & \textbf{49.82}   \\ \bottomrule
\end{tabular}
\caption{Performance on coarse and negative tasks with different backbones on the validation set of Aff-Wild2.}
\label{tab:coarse_neg}
\end{table}

\begin{table}[]
\centering
\begin{tabular}{cc}
\toprule
 $w$   & F1 (\%)   \\ \midrule
0   & 42.28     \\
32  & 43.29     \\
64  & 43.47     \\
128 & 43.69     \\
256 & \textbf{43.78}     \\
512 & 43.68    \\ \bottomrule
\end{tabular}
\caption{Evaluation of different windows size $w$ in our method on the validation set of the Aff-Wild2.}
\label{tab:window}
\end{table}

\begin{table}[]
\centering
\begin{tabular}{clc}
\toprule
Rank & Teams                      & F1 (\%)    \\ \midrule
1    & Netease Fuxi Virtual Human \cite{zhang2022transformer} & \textbf{35.87} \\
2    & IXLAB \cite{jeong2022facial} & 33.77 \\
3    & AlphaAff (Ours)                       & 32.17 \\
4    & HSE-NN \cite{savchenko2022frame}  & 30.25 \\
5    & PRL  \cite{phan2022expression}   & 28.6  \\
6    & dgu  \cite{kim2022facial}      & 27.2  \\
7    & USTC-NELSLIP                   & 21.91 \\ \midrule
     & baseline \cite{kollias2022abaw} & 20.50  \\ \bottomrule
\end{tabular}
\caption{Results of the test set of the Expression Classification Challenge.}
\label{tab:results}
\end{table}

Tab.~\ref{tab:ab} shows the ablation studies of our method on the validation set of Aff-Wild2. By using a more powerful backbone and training the whole model in an end-to-end manner, the baseline performance increased from 23 to 40. The C-F strategy brings another 2 performance increment, indicating that the C-F strategy could reduce the influence of the class imbalance problem and distinguish negative expressions efficiently. The performance further increased to 46 by applying the SP module to help the model capture both universal and unique features along the timeline.

To illustrate the impact of different backbone architectures, we show the coarse and negative results with different backbones in Fig.~\ref{fig:backbones}. As we can see, Swin-S achieves the best performance (51.52) in the coarse task and a comparable result (46.70) in the negative task, indicating the great impact achieved by the self-attention mechanism. The self-attention allows the network to use different patterns to extract different content even if they are in the same positions which are very useful to classify highly varied samples. The RepVGG achieves the best performance (49.82) in the negative task, but its performance on the coarse task is the worst (48.40). The performances on different architectures various on these two tasks, indicating they could obtain different features. Therefore, it is intuitive to ensemble them to produce robust results.

Performance with different backbone ensemble strategies is illustrated in Fig.~\ref{tab:ensemble}. The IR152 and HRNet perform badly in the negative task, so they are not considered in this experiment. The RepVGG performs worse in the coarse task, resulting in bad performance (40.38) in the overall performance. To our surprise, the Swin-S performs badly when ensemble in the negative net. When ensemble with Swin-S, the overall performance will decrease from 0.11 to 1.53. So the Swin-S, IR152, and HRNet are ensembled as the final coarse net, and the RepVGG is utilized solely as the negative net.

We also conduct experiments with different $w$ in the SP module to evaluate the impact of windows size. As is illustrated in Tab.~\ref{tab:window}, as the window size becomes larger, the model could capture more universal features, and benefit from this, it could generate smooth and stable predictions. The best result (43.78) is obtained when the size of the window is 256, which is about 8 seconds in time. When the $w$ becomes larger than 256, the performance becomes worse. This is because the expression may change in such a long period.

The final results on the test set are illustrated in Tab.~\ref{tab:results}, our method achieved 3rd place on the leader board. Different from other methods, which utilize other modalities (\eg voice, spoken words), we only adopt frame images as our input. We believe with the help of modality fusion modules proposed in other methods, the performance could increase further.

\section{Conclusion}
In this paper, we have proposed the Coarse-to-Fine Cascaded Networks with Smooth Predicting (CFC-SP) for video facial expression recognition. To address the problem of label ambiguity, we first group several similar emotions to form a coarse category. Then, the cascaded networks were employed to first conduct a coarse classification and then perform a fine classification. Moreover, we proposed a Smooth Predicting (SP) to further improve the performance. Experiments on Aff-Wild2 have verified the effectiveness of the proposed method.

{\small
\bibliographystyle{ieee_fullname}
\bibliography{egbib}
}

\end{document}